\documentclass[11pt]{article}

\usepackage[preprint]{acl}

\usepackage{times}
\usepackage{latexsym}

\usepackage[T1]{fontenc}
\usepackage[utf8]{inputenc}

\usepackage{microtype}

\usepackage{inconsolata}

\usepackage{graphicx}

\usepackage{booktabs}
\usepackage{fvextra}

\usepackage{tabularx}

\usepackage{algorithm}
\usepackage{algorithmic}

\usepackage[most]{tcolorbox}
\usepackage{ragged2e}
\usepackage{xurl}

\newtcolorbox{promptbox}[1][]{
  breakable,
  enhanced,
  width=\linewidth,
  colback=gray!5,
  colframe=gray!45,
  boxrule=0.5pt,
  arc=2pt,
  left=4pt,
  right=4pt,
  top=4pt,
  bottom=4pt,
  fonttitle=\bfseries,
  fontupper=\footnotesize,
  before upper={\RaggedRight\setlength{\parindent}{0pt}\setlength{\emergencystretch}{2em}},
  #1
}

\usepackage{multirow}

\title{Tournament-GRPO: Group-Wise Tournament Rewards for Reinforcement Learning in Open-Ended Long-Form Generation}

\author{
\textbf{Zixuan Yang}$^{1}$\thanks{Equal contribution.}, 
\textbf{Yiqun Chen}$^{1}$\footnotemark[1], 
\textbf{Wei Yang}$^{2}$\footnotemark[1], 
\textbf{Erhan Zhang}$^{1}$, 
\textbf{Zihan Shen}$^{3}$, \\
\textbf{Xiaochi Wei}$^{4}$, 
\textbf{Yan Gao}$^{4}$, 
\textbf{YIWU}$^{4}$, 
\textbf{Yao Hu}$^{4}$, 
\textbf{Jiaxin Mao}$^{1}$\thanks{Corresponding author.} \\
$^{1}$ Renmin University of China \\
$^{2}$  University of Southern California \\
$^{3}$ Zhejiang University \\
$^{4}$ Xiaohongshu Inc. \\
\texttt{zixuanyang@ruc.edu.cn}, \texttt{chenyiqun990321@ruc.edu.cn}, \texttt{maojiaxin@gmail.com}
}

\begin{document}
\maketitle

\begin{abstract}
Reinforcement learning in open-ended long-form generation is challenging because reliable reference answers and automatic metrics are often unavailable. Existing rubric-based methods typically rely on pointwise LLM-as-a-judge scoring, but absolute scores are difficult to calibrate across complex responses, may provide weak discrimination among same-query rollouts, and can become saturated during optimization. We propose \textbf{Tournament-GRPO}, a group-wise reward framework that converts rubric-guided LLM judgments into relative rewards through repeated multi-round tournaments among same-query rollouts. Tournament-GRPO compares candidates within groups, accumulates tournament outcomes, and normalizes them into group-wise rewards for GRPO training. Experiments on Deep Research Bench show that Tournament-GRPO consistently outperforms existing reward-design baselines, achieving a 4.52-point overall-score improvement over the strongest baseline. Further analyses show that tournament rewards provide a favorable effectiveness--efficiency trade-off and that tournament design affects training dynamics. These results suggest that rubric-guided tournament comparison provides an effective reward signal for reinforcement learning in open-ended long-form generation.
\footnote{Our code and pre-processing scripts are available at
\url{https://github.com/PuffYang/Tournament_GRPO}.}
\end{abstract}

\section{Introduction}

Large language models (LLMs) are increasingly used for open-ended information-seeking tasks such as deep research, report generation, and complex question answering~\cite{nakano2021webgpt,liu2023webglm}. Unlike standard question answering, where outputs can often be evaluated with exact match or token-level F1, long-form responses may synthesize evidence from multiple sources, follow task-specific rubrics, and admit many valid forms~\cite{kamalloo2023evaluating,sharma2025researchrubrics}. This makes reward design a central challenge for reinforcement learning (RL) in open-ended long-form generation.

\begin{figure}[t]
  \centering
  \includegraphics[width=0.49\textwidth]{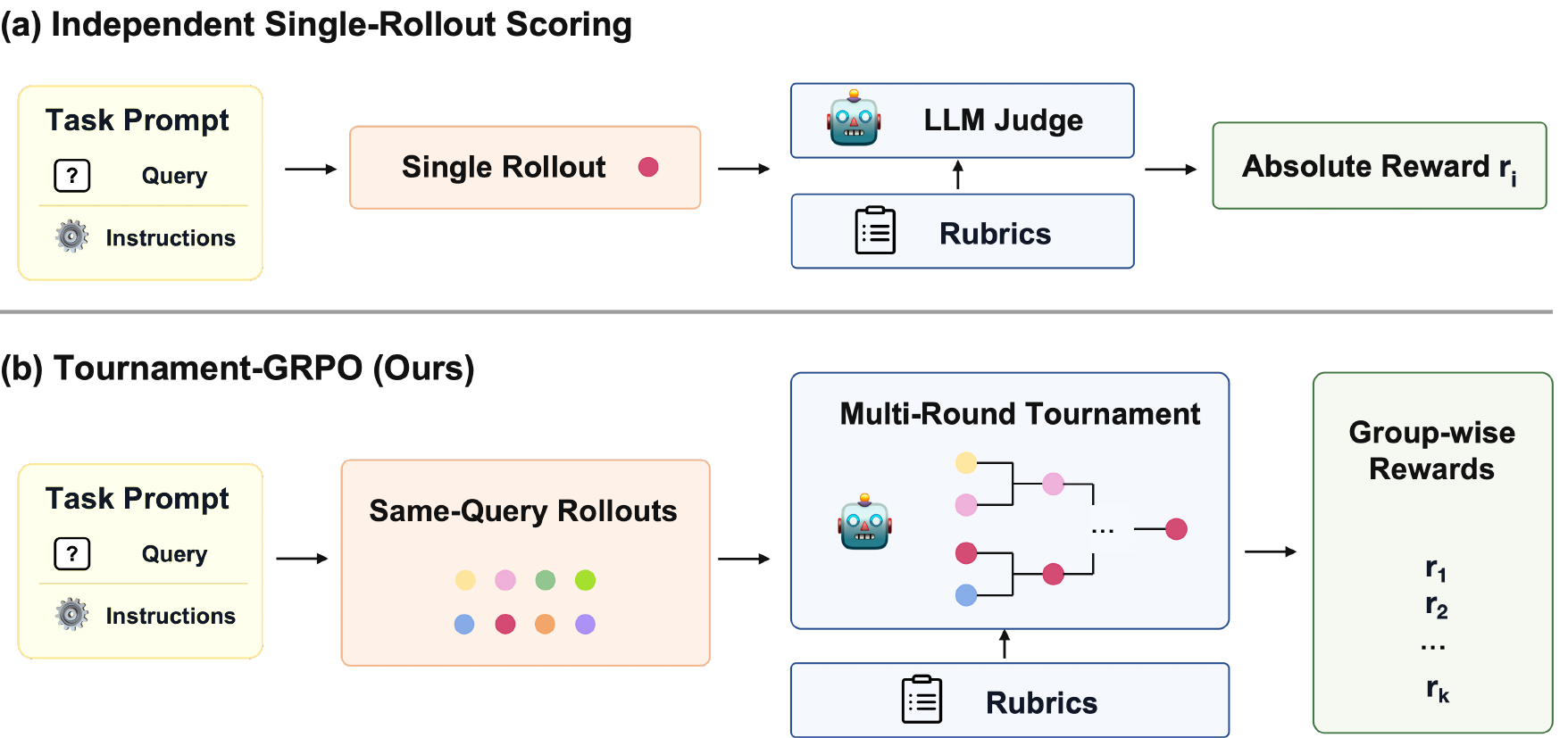}
  \caption{Comparison between independent single-rollout scoring and Tournament-GRPO. Independent scoring assigns an absolute rubric-based reward to each rollout separately, whereas Tournament-GRPO performs rubric-guided comparisons among same-query rollouts and converts tournament outcomes into normalized group-wise rewards.}
  \label{fig:comparison}
\end{figure}

A common solution is to use LLM-as-a-judge with task-specific rubrics to assign scalar rewards when reference answers or deterministic metrics are unavailable~\cite{weiqurl,gunjal2025rubrics}. In many rubric-based RL settings, each rollout is evaluated independently against a set of rubrics, and the resulting absolute score is used for policy optimization~\cite{gunjal2025rubrics,xu2026alternating}. While flexible and scalable, this pointwise absolute-scoring paradigm has several limitations for RL in open-ended long-form generation.

First, \textbf{absolute reward calibration is difficult}. A judge must map complex long-form responses, heterogeneous queries, and multiple rubric dimensions onto a single scalar scale. Such scores may be poorly calibrated across queries, rubrics, and training stages, so the same numerical reward may not correspond to the same level of response quality. Second, \textbf{pointwise rubric scores may have weak discriminative ability}. Some rubric dimensions can have much smaller variance than others, or many sampled rollouts may receive similar scores on a saturated dimension. In this case, the aggregated reward provides little useful separation among same-query rollouts, making advantage estimation noisy for group-based RL algorithms such as GRPO~\cite{shao2024deepseekmath}. Third, \textbf{absolute scoring can be vulnerable to optimization-induced saturation}. As training progresses, the policy may learn to satisfy easy or stylistic rubric dimensions, causing certain rubric scores to saturate while more subtle aspects such as evidence use, insight, or methodological rigor remain under-optimized. This can reduce the usefulness of absolute rewards and may encourage reward hacking along the dimensions that are easiest for the judge to score.

These issues are especially problematic because GRPO optimizes a policy using multiple rollouts sampled from the same query. In this setting, the reward does not only need to be a plausible standalone quality score; it must reliably distinguish which rollout is better within the same-query group. Recent work on LLM-as-a-judge evaluation also suggests that feedback protocols matter: LLM judges can be sensitive to bias and instability~\cite{shi2025judging}, and relative comparison can differ substantially from pointwise scoring in reliability and preference consistency~\cite{tripathi2025pairwise,gao2025re}. This motivates a reward design that aligns more directly with the group-relative structure of GRPO.

Motivated by TourRank~\cite{chen2025tourrank}, which ranks candidates through iterative within-group comparisons, we propose \textbf{Tournament-GRPO}, a group-wise tournament reward framework that converts rubric-guided LLM judgments into relative rewards for GRPO optimization. Rather than scoring each rollout independently, Tournament-GRPO groups same-query rollouts and conducts repeated multi-round tournaments, where shuffled rollouts are partitioned into comparison groups and judged according to the query, rubrics, and candidate responses. Winners receive points and advance across rounds, and the accumulated points are normalized within each query group to form a relative \emph{tournament reward}. This design directly addresses the limitations of pointwise absolute scoring: rubric-guided comparisons reduce reliance on globally calibrated scalar scores, ranking-based tournament outcomes preserve separation among same-query rollouts, and group-wise normalization makes the reward better aligned with GRPO's relative advantage estimation. Figure~\ref{fig:comparison} illustrates the key difference between independent single-rollout scoring and our group-wise tournament reward construction.

Our contributions are summarized as follows:
\begin{itemize}
    \item We identify key limitations of pointwise rubric-based reward design for RL in open-ended long-form generation, including poor absolute-score calibration, limited discriminative ability among same-query rollouts, and optimization-induced reward saturation.

    \item We propose \textbf{Tournament-GRPO}, a group-wise tournament reward framework that compares same-query rollouts through repeated rubric-guided tournaments and converts accumulated tournament outcomes into normalized rewards for GRPO training.

    \item We conduct experiments on Deep Research Bench~\cite{du2025deepresearch} against multiple reward-design baselines. Tournament-GRPO achieves the best overall performance, provides a better effectiveness--efficiency trade-off than exhaustive pairwise comparison, and remains effective under additional backbone and judge settings.
\end{itemize}

\begin{figure*}[t]
  \centering
  \includegraphics[width=1.0\textwidth]{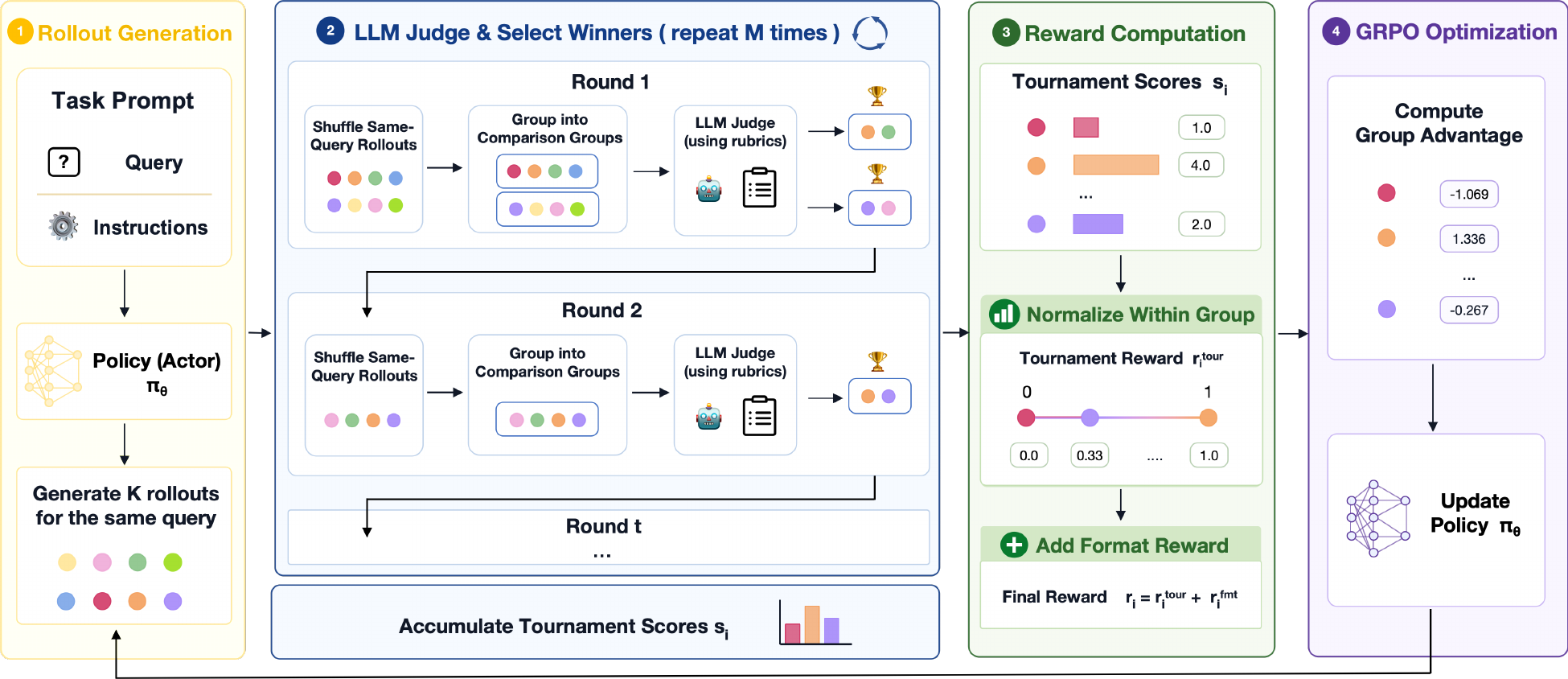}
  \caption{Overview of Tournament-GRPO. For each query, we sample K rollouts and perform a multi-round tournament over M repeats. In each round, rollouts are grouped, judged by an LLM using the rubrics, and winners advance. Winners accumulate points across rounds and repeats. The points are normalized within the query group to produce group-wise rewards, combined with a format reward, and used by GRPO to compute group advantages and update the policy.
}
  \label{fig:framework}
\end{figure*}

\section{Related Work}

\paragraph{RL for LLMs.}
RL has become a widely used approach for aligning LLMs with human preferences and task-specific objectives. Early preference-based RL learns reward functions from human comparisons~\cite{christiano2017deep}, and RLHF has been widely adopted for language-model alignment~\cite{stiennon2020learning,ouyang2022training}. These methods commonly optimize policies with PPO-style algorithms~\cite{schulman2017proximal}, while later work further explores learning from AI feedback or principle-based feedback~\cite{bai2022constitutional} and direct preference optimization without an explicit RL loop~\cite{rafailov2023direct}.

More recently, RL with rule-based or verifiable rewards has been used to improve reasoning and problem-solving abilities in LLMs~\cite{guo2025deepseek,wen2025reinforcement}. In particular, GRPO estimates advantages by comparing multiple sampled rollouts within the same prompt group~\cite{shao2024deepseekmath,guo2025deepseek}. Our work follows this group-based RL setting, but targets open-ended long-form generation, where such deterministic reward functions are typically unavailable and evaluation must rely on rubric-based semantic judgments.

\paragraph{LLM-as-a-judge and rubric-based evaluation.}
For open-ended generation tasks, reference-based automatic metrics often fail to capture factuality, coherence, and instruction satisfaction. LLM-as-a-judge has therefore become a common evaluation paradigm, where a strong language model assesses generated responses according to task-specific rubrics~\cite{zheng2023judging,liu2023g,li2023alpacaeval}. Recent benchmarks further study the reliability of reward models and LLM judges~\cite{lambert2025rewardbench,tan2025judgebench}.
Rubric-based evaluators further provide fine-grained feedback and scores for different aspects of response quality~\cite{kim2024prometheus}. However, LLM-as-a-judge is also known to suffer from biases and instability, including position bias, verbosity bias, and sensitivity to response length~\cite{zheng2023judging,dubois2024length}. These issues can become problematic when raw judge scores are directly used as scalar rewards for RL.

\paragraph{Relative comparison and tournament ranking.}
Compared with assigning calibrated absolute scores, relative comparison is often a more natural and reliable judgment format. Preference-based alignment methods use comparative feedback between candidate outputs~\cite{stiennon2020learning,bai2022constitutional}. In ranking tasks, TourRank~\cite{chen2025tourrank} proposes a tournament-inspired strategy that uses LLMs to compare candidate documents across multiple rounds and obtain more robust rankings. These studies suggest that relative judgments can provide more stable signals than independent absolute scoring; our work extends this insight from inference-time ranking to group-wise reward construction for RL over same-query rollouts.

\section{Method}
\label{sec:method}
As shown in Figure~\ref{fig:framework}, Tournament-GRPO compares multiple rollouts sampled from the same query through repeated multi-round tournaments, and converts the resulting tournament outcomes into group-wise rewards for GRPO training.

\subsection{Tournament Reward}

Given a query $q$ and a set of task-specific rubrics $\mathcal{B}$, the policy $\pi_\theta$ samples $K$ rollouts:
\begin{equation}
    \mathcal{Y}_q = \{y_1,\ldots,y_K\}, 
    \quad y_i \sim \pi_\theta(\cdot \mid q)
\end{equation}

For each query, Tournament-GRPO repeats the multi-round tournament process $M$ times among same-query rollouts. Let $\mathbf{s}=[s_1,\ldots,s_K]$ denote the accumulated tournament scores, initialized as $s_i=0$. At the beginning of each repeat, all same-query rollouts are treated as active candidates:
\begin{equation}
    \mathcal{C}^{(0)} = \{y_1,\ldots,y_K\}
\end{equation}

At each round $t$, the current active candidate set $C^{(t)}$ is first
shuffled to reduce order bias and then partitioned into comparison groups
$\{P_1^{(t)},\ldots,P_L^{(t)}\}$, where each group contains $G$ candidates.
For simplicity, our experiments use configurations in which $|C^{(t)}|$ is
divisible by the group size $G$. These groups form a disjoint
partition of $C^{(t)}$:
\begin{equation}
    \bigcup_{l=1}^{L} P_l^{(t)} = \mathcal{C}^{(t)}, 
    \quad
    P_l^{(t)} \cap P_{l'}^{(t)} = \emptyset \ (l \neq l')
\end{equation}

For each group, an LLM judge $J_{\mathrm{tournament}}$ selects exactly $K_{\mathrm{win}}$ winners to advance to the next round:
\begin{equation}
    W_l^{(t)}
    =
    J_{\mathrm{tournament}}\left(q,\mathcal{B},\{y_i:i\in P_l^{(t)}\}\right)
\end{equation}
where $W_l^{(t)} \subseteq P_l^{(t)}$ denotes the selected winners and
$|W_l^{(t)}| = K_{\mathrm{win}}$ for each group $l$.
Winners receive points $\omega_t$ and form the active candidates for the next round:
\begin{equation}
    s_i \leftarrow s_i + \omega_t ,
    \quad \forall i \in W_l^{(t)}
\end{equation}
\begin{equation}
    \mathcal{C}^{(t+1)} = \bigcup_{l=1}^{L} W_l^{(t)}
\end{equation}

The tournament stops when the number of winners after a round is no larger than a final-candidate threshold $K_{\mathrm{final}}$:
\begin{equation}
    \sum_{l=1}^{L} |W_l^{(t)}| \leq K_{\mathrm{final}}
\end{equation}

\subsection{Reward Normalization and GRPO Integration}
\label{sec:reward_normalization}
After repeating the tournament process $M$ times, each rollout obtains an accumulated score $s_i$. We normalize these scores within the same-query group to obtain the tournament reward $r_i^{\mathrm{tour}}$. We use min-max normalization because it can preserve the relative order of rollouts within the group:
\begin{equation}
    r_i^{\mathrm{tour}}
    =
    \frac{s_i-\min_j s_j}{\max_j s_j-\min_j s_j+\epsilon}
\end{equation}

We also use a binary format reward $r_i^{\mathrm{fmt}}\in\{0,1\}$ to enforce the required format. A rollout receives $r_i^{\mathrm{fmt}}=1$ only if it follows the expected format.

The final training reward is the sum of the tournament reward and the format reward:
\begin{equation}
    r_i = r_i^{\mathrm{tour}} + r_i^{\mathrm{fmt}}.
\end{equation}
This reward is then used by GRPO to compute group-relative advantages:
\begin{equation}
    A_i =
    \frac{r_i-\mu(\mathbf{r})}{\sigma(\mathbf{r})+\epsilon}
\end{equation}
where $\mathbf{r}=[r_1,\ldots,r_K]$ contains rewards for rollouts sampled from the same query. Because $r_i^{\mathrm{tour}}$ is derived from relative comparisons within the same-query group, it naturally matches the group-relative structure of GRPO.

The overall tournament reward computation procedure is summarized in Algorithm~\ref{alg:tournament_grpo}.

\begin{algorithm}[t]
\small
\caption{Tournament-GRPO Reward Computation}
\label{alg:tournament_grpo}
\begin{algorithmic}[1]
\REQUIRE Query $q$, rubrics $\mathcal{B}$, rollouts $\{y_i\}_{i=1}^{K}$, the number of tournament repeats $M$, group size $G$, winners per group $K_{\mathrm{win}}$, final threshold $K_{\mathrm{final}}$, judge $J_{\mathrm{tournament}}$
\ENSURE Tournament rewards $\{r_i^{\mathrm{tour}}\}_{i=1}^{K}$

\STATE Initialize tournament scores $s_i \leftarrow 0$ for all $i$
\FOR{$m=1$ to $M$}
    \STATE $\mathcal{C}^{(0)} \leftarrow \{1,\ldots,K\}$
    \STATE $t \leftarrow 0$
    \WHILE{$|\mathcal{C}^{(t)}| > K_{\mathrm{final}}$}
        \STATE $\widetilde{\mathcal{C}}^{(t)} \leftarrow \mathrm{Shuffle}(\mathcal{C}^{(t)})$
        \STATE Partition $\widetilde{\mathcal{C}}^{(t)}$ into groups $\{P_l^{(t)}\}_{l=1}^{L}$ of size $G$
        \STATE $\mathcal{C}^{(t+1)} \leftarrow \emptyset$
        \FOR{$l=1$ to $L$}
            \STATE $W_l^{(t)} \leftarrow J_{\mathrm{tournament}}\!\left(q,\mathcal{B},\{y_i:i\in P_l^{(t)}\}\right)$
            \FOR{$i \in W_l^{(t)}$}
                \STATE $s_i \leftarrow s_i + \omega_t $
            \ENDFOR
            \STATE $\mathcal{C}^{(t+1)} \leftarrow \mathcal{C}^{(t+1)} \cup W_l^{(t)}$
        \ENDFOR
        \STATE $t \leftarrow t+1$
    \ENDWHILE
\ENDFOR

\STATE Normalize scores within the query group:
\[
r_i^{\mathrm{tour}} =
\frac{s_i-\min_j s_j}{\max_j s_j-\min_j s_j+\epsilon}
\]
\RETURN $\{r_i^{\mathrm{tour}}\}_{i=1}^{K}$
\end{algorithmic}
\end{algorithm}

\section{Experiments}
\label{sec:experiments}

We evaluate Tournament-GRPO on Deep Research Bench~\cite{du2025deepresearch}, a benchmark for open-ended long-form research tasks. Our experiments are designed to answer the following questions:
(1) whether group-wise tournament rewards improve GRPO training compared with independent absolute rubric scoring;
(2) how the number of tournament repeats $M$ affects performance; and
(3) how different tournament structures, including comparison group size $G$ and the number of winners per group $K_{\mathrm{win}}$, influence reward quality and training stability.

\subsection{Experimental Setup}
\label{sec:experimental_setup}

\paragraph{Training framework and backbone.}
We build our RL training pipeline on top of \texttt{verl} \footnote{\url{https://github.com/volcengine/verl}} and optimize the policy with GRPO. The policy backbone is Qwen2.5-7B-Instruct~\cite{qwen2025qwen25technicalreport}. For judge-based reward computation, we use a self-deployed Qwen2.5-72B-Instruct~\cite{qwen2025qwen25technicalreport} as the LLM judge. 

\paragraph{Training data.}
For our RL training, we use the training data from the RL stage of DR Tulu~\cite{shao2025dr}. The dataset contains prompts that require open-ended long-form answer generation. We split the data into 95\% training and 5\% validation sets for our experiments.

\paragraph{Query-specific rubrics.}
Each training query is paired with a set of query-specific rubrics. We generate these rubrics with GPT-4o before RL training and keep them fixed throughout training. The full rubric-generation prompt is provided in Appendix~\ref{app:rubric_generation_prompt}.

\paragraph{Tools and rollout format.}
Following the ReAct paradigm of tool-augmented language agents~\cite{yao2022react}, we implement two tools in the environment: \texttt{google\_search}, which returns web search results, and \texttt{browse\_webpage}, which reads the content of a selected URL.
During rollout, the policy samples 8 rollouts per prompt. We provide the full policy prompt in Appendix~\ref{app:policy_prompt}. The main implementation hyperparameters are reported in Appendix~\ref{app:implementation_details}.

\paragraph{Tournament-GRPO Configurations.}
We evaluate two tournament structures:
$(G=2,K_{\mathrm{win}}=1,K_{\mathrm{final}}=1)$ and
$(G=4,K_{\mathrm{win}}=2,K_{\mathrm{final}}=2)$. For each structure,
we vary the number of tournament repeats $M$ to study how repeated comparisons
affect reward quality and training performance. A non-uniform $\omega_t$ could emphasize later-round wins, but it also introduces an additional design choice that may interact with $G$, $K_{\mathrm{win}}$, $K_{\mathrm{final}}$, and $M$. We therefore
use $\omega_t=1$ as a simple default and tune only the tournament structure. The full tournament judge prompt is provided in Appendix~\ref{app:tournament_judge_prompt}.

\subsection{Baselines}
\label{sec:baselines}
The baseline prompts and reward computation details are provided in Appendices~\ref{app:baseline_judge_prompts} and~\ref{app:baseline_reward_computation}.
\footnote{
Unless otherwise specified, all RL methods use the same policy backbone, training data, rollout budget, tool-use setting, format reward, judge model, and training steps; they differ only in how judge feedback is converted into scalar rewards.}

\paragraph{Zero-shot.}
We evaluate the base policy without RL training to measure its initial performance.

\paragraph{Implicit and explicit absolute scoring.}
We consider GRPO~\cite{shao2024deepseekmath} and Dr.GRPO~\cite{liu2025understanding} variants with independent judge scores. The implicit variants ask the judge to assign a single scalar score to each rollout, while the explicit variants compute rewards from rubric-level scores. GDPO~\cite{liu2026gdpo} is included only in the explicit setting. We do not include an implicit GDPO variant because, when the reward is a single scalar judge score, GDPO reduces to the same reward formulation as GRPO implicit, apart from the shared format reward.

\paragraph{Pairwise comparison.}
Motivated by the pairwise comparison strategy in ArenaRL~\cite{zhang2026arenarl},
we include a relative-comparison baseline that exhaustively compares same-query
rollouts in pairs. For each rollout, we compute its win rate over all
pairwise comparisons and use this win rate as the scalar reward for GRPO
training. This baseline isolates the effect of replacing independent absolute
scoring with direct relative comparison.

\subsection{Main Results}
\label{sec:main_results}

\begin{table*}[t]
\centering
\small
\begin{tabular}{llccccc}
\hline
\textbf{Group} & \textbf{Method} & \textbf{Readability} & \textbf{Instruct.} & \textbf{Compre.} & \textbf{Insight} & \textbf{Overall} \\
\hline
\multirow{7}{*}{Baseline}
& Zero-shot & 35.40 & 34.21 & 31.42 & 29.76 & 32.13 \\
& Dr. GRPO implicit & 38.93 & 37.78 & 34.51 & 34.55 & 35.95 \\
& Dr. GRPO explicit & 40.03 & 38.47 & 35.80 & 35.54 & 36.97 \\
& GDPO & 43.22 & 42.07 & 39.81 & 39.05 & 40.61 \\
& GRPO explicit & 41.75 & 43.43 & 43.04 & 42.61 & 42.79 \\
& GRPO implicit & 41.41 & 40.23 & 37.85 & 36.88 & 38.60 \\
& Pairwise & 47.01 & 47.25 & 46.36 & 45.22 & 46.25 \\
\hline
\multirow{5}{*}{$G=2$, $K_{\mathrm{win}}=1$, $K_{\mathrm{final}}=1$}
& $M=1$ & 43.81 & 43.58 & 42.40 & 40.98 & 42.40 \\
& $M=2$ & \underline{51.39} & \underline{51.51} & \underline{50.69} & \underline{50.82} & \underline{51.01} \\
& $M=3$ & \textbf{52.09} & \textbf{52.71} & \textbf{52.32} & \textbf{52.04} & \textbf{52.28} \\
& $M=4$ & 46.92 & 47.59 & 46.50 & 45.02 & 46.33 \\
& $M=8$ & 47.90 & 47.91 & 47.11 & 45.84 & 46.98 \\
\hline
\multirow{5}{*}{$G=4$, $K_{\mathrm{win}}=2$, $K_{\mathrm{final}}=2$}
& $M=1$ & 49.70 & 49.53 & 48.69 & 48.02 & 48.79 \\
& $M=2$ & 50.08 & 50.32 & 49.90 & 49.82 & 49.99 \\
& $M=3$ & 43.00 & 43.32 & 42.78 & 41.84 & 42.60 \\
& $M=4$ & 50.31 & 50.34 & 49.88 & 49.49 & 49.90 \\
& $M=8$ & 41.48 & 42.03 & 41.28 & 40.39 & 41.17 \\
\hline
\end{tabular}
\caption{Results on Deep Research Bench after the first training epoch at step 600. Evaluation uses a self-deployed Qwen2.5-72B-Instruct as the LLM judge.}
\label{tab:epoch1_results}
\end{table*}

\begin{table*}[t]
\centering
\small
\begin{tabular}{llccccc}
\hline
\textbf{Group} & \textbf{Method} & \textbf{Readability} & \textbf{Instruct.} & \textbf{Compre.} & \textbf{Insight} & \textbf{Overall} \\
\hline
\multirow{6}{*}{Baseline}
& Dr. GRPO implicit & 40.88 & 38.92 & 36.42 & 35.47 & 37.38 \\
& Dr. GRPO explicit & 40.49 & 38.98 & 36.40 & 36.07 & 37.51 \\
& GDPO & 51.11 & 51.01 & 50.52 & 50.52 & 50.71 \\
& GRPO explicit & 41.26 & 41.56 & 41.33 & 40.77 & 41.20 \\
& GRPO implicit & 33.09 & 32.97 & 32.14 & 31.17 & 32.14 \\
& Pairwise & 51.15 & 51.34 & 50.38 & 50.05 & 50.57 \\
\hline
\multirow{5}{*}{$G=2$, $K_{\mathrm{win}}=1$, $K_{\mathrm{final}}=1$}
& $M=1$ & 50.06 & 50.76 & 50.16 & 49.94 & 50.20 \\
& $M=2$ & 52.38 & 53.04 & 53.17 & 52.91 & 52.96 \\
& $M=3$ & 48.69 & 49.43 & 47.94 & 46.24 & 47.81 \\
& $M=4$ & \underline{53.69} & \underline{54.92} & \textbf{55.63} & \underline{55.16} & \underline{55.04} \\
& $M=8$ & 52.93 & 53.90 & 53.96 & 54.04 & 53.84 \\
\hline
\multirow{5}{*}{$G=4$, $K_{\mathrm{win}}=2$, $K_{\mathrm{final}}=2$}
& $M=1$ & 52.79 & 53.26 & 52.95 & 53.38 & 53.14 \\
& $M=2$ & 53.03 & 53.54 & 53.83 & 53.80 & 53.64 \\
& $M=3$ & 45.68 & 46.00 & 46.23 & 46.01 & 46.04 \\
& $M=4$ & 49.65 & 50.92 & 50.53 & 50.01 & 50.32 \\
& $M=8$ & \textbf{53.81} & \textbf{55.22} & \underline{55.19} & \textbf{55.40} & \textbf{55.09} \\
\hline
\end{tabular}
\caption{Results on Deep Research Bench after the second training epoch at step 1100. Evaluation uses a self-deployed Qwen2.5-72B-Instruct as the LLM judge.}
\label{tab:epoch2_results}
\end{table*}

Tournament rewards are inherently relative: they measure how a rollout performs against other sampled rollouts for the same query, rather than its absolute task quality. Therefore, the tournament reward curve cannot be used as a reliable convergence indicator, since the comparison set changes with sampling, grouping, and policy updates. We use training rewards only to monitor whether reward computation behaves normally, and rely on out-of-domain evaluation to assess model performance and analyze tournament hyperparameters.

\begin{table*}[t]
\centering
\small
\setlength{\tabcolsep}{4pt}
\begin{tabular}{lccccc}
\hline
\textbf{Method} & \textbf{Judge-call complexity} & \textbf{Judge calls per query} & \textbf{Judge calls per rollout} & \textbf{Wall-clock} & \textbf{Overall} \\
\hline
Pointwise & $O(K) / O(KR)$ & 8 &
1.000 & 11.48 h & 38.98 \\
Pairwise & $O(K^2)$ & 28 & 3.500 & 88.59 h & 46.25 \\
Tournament, $M=1$ & $O(MK)$ & 7 & 0.875 & 59.65 h & 42.40 \\
Tournament, $M=2$ & $O(MK)$ & 14 & 1.750 & 62.91 h & 51.01 \\
Tournament, $M=3$ & $O(MK)$ & 21 & 2.625 & 67.77 h & \textbf{52.28} \\
Tournament, $M=4$ & $O(MK)$ & 28 & 3.500 & 73.05 h & 46.33 \\
\hline
\end{tabular}
\caption{Effectiveness--efficiency comparison of reward designs under the Table 1 setting with $G=2$, $K_{\mathrm{win}}=1$, and $K_{\mathrm{final}}=1$. ``Pointwise'' averages Dr. GRPO implicit $\&$ explicit, GDPO and GRPO explicit $\&$ implicit.}
\label{tab:reward_complexity}
\end{table*}

Tables~\ref{tab:epoch1_results} and~\ref{tab:epoch2_results} report the results after the first and second training epochs, respectively.

After the first epoch, pairwise comparison is the strongest baseline, reaching an overall score of 46.25. Tournament-GRPO further improves over it with multiple settings. The best first-epoch result is obtained by the binary tournament configuration $(G=2,K_{\mathrm{win}}=1,K_{\mathrm{final}}=1)$ with $M=3$, which achieves 52.28 overall, outperforming pairwise comparison by 6.03 points. This shows that group-wise tournament rewards can provide a strong early-stage learning signal while avoiding exhaustive pairwise comparisons.

After the second epoch, the advantage of Tournament-GRPO becomes more pronounced. The best configuration is $(G=4,K_{\mathrm{win}}=2,K_{\mathrm{final}}=2)$ with $M=8$, which achieves the highest overall score of 55.09. The binary tournament with $M=4$ is also competitive, reaching 55.04 overall and obtaining the best comprehensiveness score of 55.63. These results suggest that both binary and larger-group tournaments can be effective, but they favor different repeat settings and training stages.

\subsection{Optimization Dynamics across Training Epochs}
\label{sec:optimization_dynamics}

The two checkpoints show that tournament design affects both optimization speed and final performance. Different configurations are favored at different training stages, suggesting that the strength of the reward signal depends on the evolving quality distribution of same-query rollouts.

The binary tournament $(G=2,K_{\mathrm{win}}=1,K_{\mathrm{final}}=1)$ is most effective in the early stage. Its best setting, $M=3$, achieves the highest first-epoch overall score. Early in training, rollout quality is often heterogeneous: some rollouts follow the required format and cover relevant evidence, while others deviate from the task or miss key information. Binary elimination can therefore provide a sharp preference signal by directly filtering out clearly weaker rollouts.

In contrast, the larger-group tournament $(G=4,K_{\mathrm{win}}=2,K_{\mathrm{final}}=2)$ becomes more effective after longer training. Although its $M=8$ setting is weak after the first epoch, it achieves the best overall score after the second epoch. As optimization progresses, low-quality rollouts become less frequent and same-query rollouts may differ in finer-grained aspects such as coverage, evidence use, and instruction following. Retaining multiple winners in a larger group can provide a softer signal that still favors stronger rollouts while avoiding overly aggressive elimination.

Overall, these results suggest that tournament structure interacts with the training stage. Sharper binary comparisons are useful when rollout quality is highly variable, whereas larger-group comparisons with multiple retained winners may be more beneficial once candidate rollouts become more competitive.

\subsection{Reward Design: Effectiveness and Efficiency}
\label{sec:reward_design_analysis}

We compare reward designs in terms of effectiveness and judge-call complexity. Table~\ref{tab:reward_complexity} additionally reports measured judge calls and training wall-clock time. Here, $R$ denotes the number of rubrics; the remaining notation is defined in Section~\ref{sec:method}.

Pointwise reward design is efficient but less effective: implicit and explicit scoring require $O(K)$ and $O(KR)$
judge calls, respectively, but achieve an average overall score of only 38.98. Because they score each rollout
independently, the judge cannot observe the quality of other rollouts in the same group, making it difficult to assign well-calibrated rewards within a group. 
Pairwise comparison avoids absolute-score calibration and reaches 46.25 overall, but it relies on brute-force enumeration of all pairs, resulting in an $O(K^2)$ cost.
Tournament-GRPO retains relative comparison without full pairwise enumeration. Starting from $K$ rollouts, each round groups candidates in sets of size $G$ and retains $K_{\mathrm{win}}$ winners. Thus, candidates decrease by approximately $K_{\mathrm{win}}/G$ per round, and the number of comparison groups at round $t$ is approximately
$\frac{K}{G}\left(\frac{K_{\mathrm{win}}}{G}\right)^{t-1}$.
The tournament stops when the number of remaining candidates reaches $K_{\mathrm{final}}$, which requires approximately
$\left\lceil \log_{G/K_{\mathrm{win}}} \frac{K}{K_{\mathrm{final}}} \right\rceil$
rounds. Therefore, the total judge-call cost is proportional to:
\[
M \sum_{t=1}^{\left\lceil \log_{G/K_{\mathrm{win}}} \frac{K}{K_{\mathrm{final}}} \right\rceil}
\frac{K}{G}
\left(\frac{K_{\mathrm{win}}}{G}\right)^{t-1}
\]
Since $K_{\mathrm{win}} < G$, this is a geometric series and is linear in $K$ up to a constant factor determined by $G$, $K_{\mathrm{win}}$, and $K_{\mathrm{final}}$. Hence, the asymptotic judge-call complexity of Tournament-GRPO is $O(MK)$.

Compared with exhaustive pairwise comparison, Tournament-GRPO with $M=3$ uses 25\% fewer judge calls (21 versus 28 per query), reduces wall-clock time by 23.5\% (67.77 versus 88.59 hours), and improves the score by 6.03 points. The $M=2$ configuration is even cheaper while still improving performance by 4.76 points. This suggests that tournament-based reward aggregation offers a favorable effectiveness--efficiency trade-off when learning from rubric-based LLM-judge feedback.

\subsection{Tournament Hyperparameter Analysis}
\label{sec:tournament_hyperparameter_analysis}
Tournament performance is affected by multiple design choices, including the
comparison group size $G$, winner count $K_{\mathrm{win}}$, final-candidate
threshold $K_{\mathrm{final}}$, and repeat count $M$. Tables~\ref{tab:epoch1_results}
and~\ref{tab:epoch2_results} show that the best repeat count varies across
tournament structures and training stages, rather than increasing monotonically
with $M$. This suggests that tournament hyperparameters interact with the
evolving policy, instead of acting as independent or uniformly beneficial
controls.

\begin{figure}[t]
\centering
\includegraphics[width=\columnwidth]{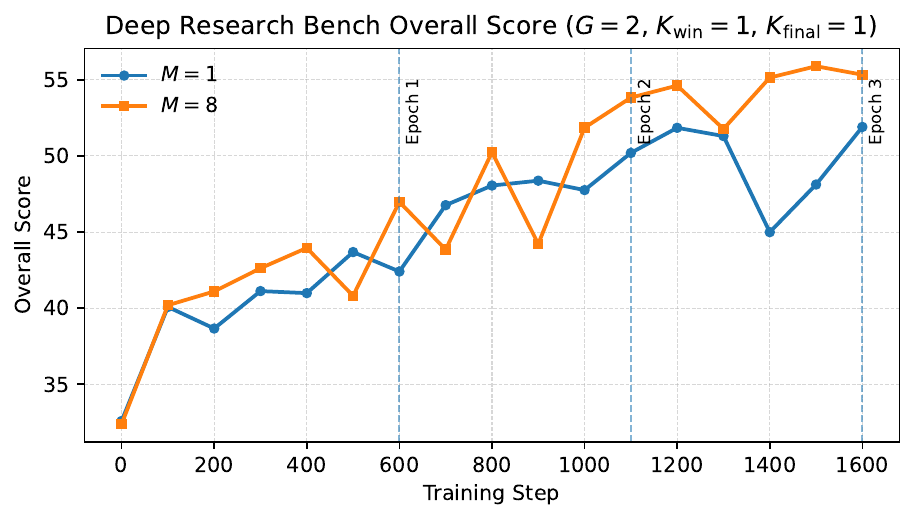}
\caption{Overall score on Deep Research Bench across training steps for Tournament-GRPO with $G=2$, $K_{\mathrm{win}}=1$, and $K_{\mathrm{final}}=1$. We compare $M=1$ and $M=8$.}
\label{fig:m1_m8_training_curve}
\end{figure}

Figure~\ref{fig:m1_m8_training_curve} provides a dynamic view of the effect of tournament repeats. Both $M=1$ and $M=8$ improve substantially over training, but they follow different trajectories. The $M=1$ setting improves more steadily and reaches 51.91 after the third epoch. In contrast, $M=8$ is less monotonic during training but achieves stronger final performance, reaching 55.34 at step 1600. This suggests that more tournament repeats can provide a stronger final reward signal, although they may also lead to less stable intermediate dynamics.

\begin{figure}[t]
\centering
\includegraphics[width=\columnwidth]{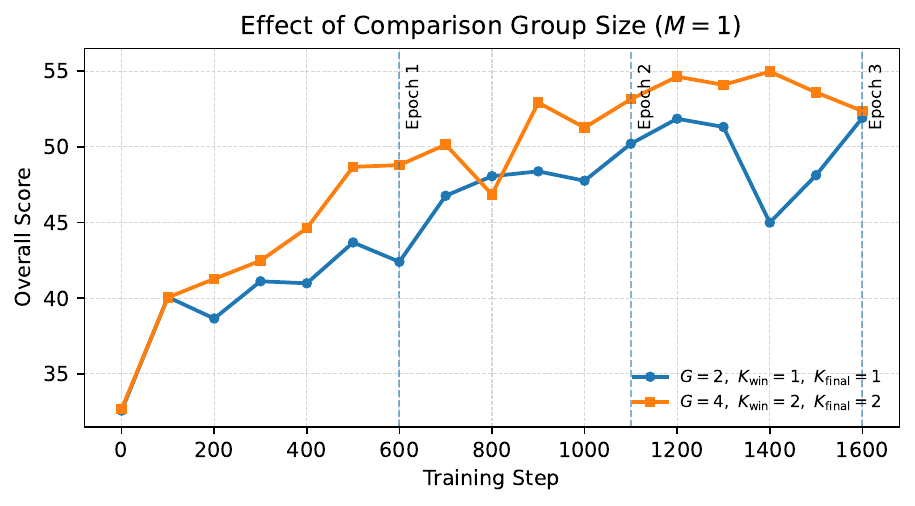}
\caption{Overall score on Deep Research Bench across training steps for Tournament-GRPO with $M=1$. We compare $(G=2,K_{\mathrm{win}}=1,K_{\mathrm{final}}=1)$ and $(G=4,K_{\mathrm{win}}=2,K_{\mathrm{final}}=2)$.}
\label{fig:group_size_repeat1_curve}
\end{figure}

\begin{figure}[t]
\centering
\includegraphics[width=\columnwidth]{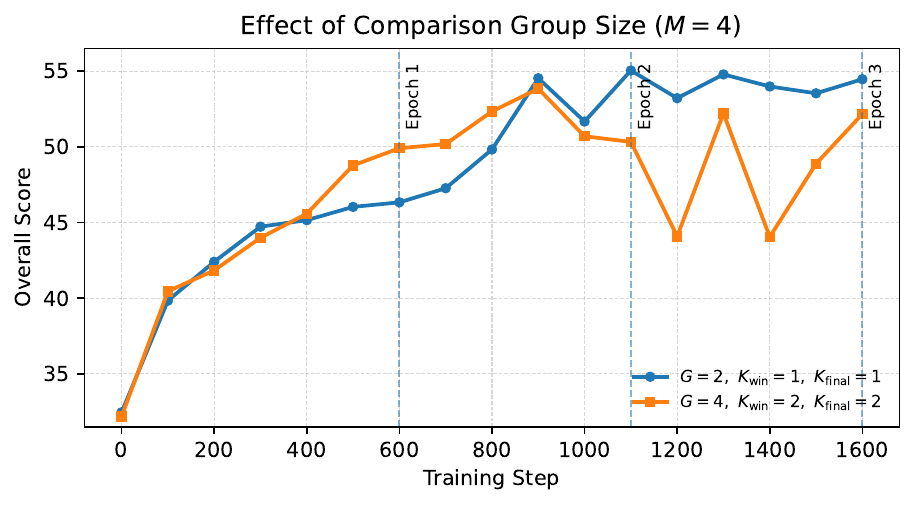}
\caption{Overall score on Deep Research Bench across training steps for Tournament-GRPO with $M=4$. We compare $(G=2,K_{\mathrm{win}}=1,K_{\mathrm{final}}=1)$ and $(G=4,K_{\mathrm{win}}=2,K_{\mathrm{final}}=2)$.}
\label{fig:group_size_repeat4_curve}
\end{figure}

Figures~\ref{fig:group_size_repeat1_curve} and~\ref{fig:group_size_repeat4_curve}
compare different tournament structures under fixed values of the repeat count
$M$. When $M=1$, the larger-group setting $(G=4,K_{\mathrm{win}}=2,K_{\mathrm{final}}=2)$ outperforms the smaller-group setting across most training steps, reaching 53.14 versus 50.20 at step 1100. This suggests that larger comparison groups can provide stronger local comparison signals when only one tournament repeat is used.

When $M=4$, the trend changes. The smaller-group setting $(G=2,K_{\mathrm{win}}=1,K_{\mathrm{final}}=1)$ becomes stronger after sufficient training, reaching 55.04 at step 1100 and 54.46 at step 1600, compared with 50.32 and 52.16 for the larger-group setting. This indicates that the benefit of a larger comparison group is not universal; when repeated tournaments are used, smaller pairwise-like comparisons can yield sharper and more stable preference signals.

Overall, these results show that tournament design involves a trade-off between comparison richness, comparison sharpness, and training dynamics. Larger groups expose the judge to more candidates within each local decision, while smaller groups simplify the judgment and may work better when tournament outcomes are aggregated across multiple repeats. Therefore, $M$, $G$, $K_{\mathrm{win}}$, and $K_{\mathrm{final}}$ should be tuned jointly. 

Appendix~\ref{app:additional_backbone_judge_results} presents additional robustness and generalization results, while Appendix~\ref{app:case_study} provides a case study.

\section{Conclusion}
\label{sec:conclusion}

We presented \textbf{Tournament-GRPO}, a group-wise reward framework for RL in open-ended long-form generation. Instead of relying on independently calibrated absolute scores, Tournament-GRPO performs repeated rubric-guided comparisons among same-query rollouts and converts tournament outcomes into normalized rewards for GRPO training. This better matches the group-relative nature of GRPO, where rewards should distinguish same-query rollouts rather than provide globally calibrated standalone scores. Experiments on Deep Research Bench show that Tournament-GRPO outperforms the evaluated reward-design baselines while avoiding exhaustive pairwise judging. Further analyses demonstrate its effectiveness--efficiency trade-off, the impact of tournament hyperparameters on training dynamics, and its robustness under additional backbone and judge settings.
\clearpage

\section*{Limitations}
Tournament-GRPO still relies on LLM judges, whose preferences may be biased, inconsistent, or sensitive to prompt design. Although tournament comparison reduces the need for globally calibrated absolute scores, it does not eliminate judge-side errors. In addition, tournament reward computation introduces extra judge-call overhead compared with pointwise scoring, and its performance depends on hyperparameters such as group size, winner count, final-candidate threshold, and repeat number. Finally, tournament rewards mainly provide relative preference signals among same-query rollouts; they do not directly identify specific factual errors, missing evidence, or reasoning flaws within a response. Combining tournament rewards with more fine-grained feedback or factuality verification remains a promising direction.

\section*{Ethical Considerations}
This work uses publicly available data and does not require collecting personally identifiable information. However, models trained with LLM-judge rewards may still produce factual errors, unsupported claims, or biased content, and judge-model biases may be propagated into the trained policy. Users should exercise caution when applying the proposed framework to real-world settings, especially in domains where misinformation or biased content could cause harm. We recommend using factuality checks, citation verification, and human oversight for high-stakes applications.


\bibliography{custom}

\clearpage
\appendix

\section{Rubric Generation Prompt}
\label{app:rubric_generation_prompt}
 
Before RL training, we generate query-specific rubrics for each training query. The generated rubrics are fixed and used by both Tournament-GRPO and baselines.

\begin{promptbox}[title=Rubric Generation Prompt]
Create a query-specific evaluation rubric for the user query below.\\

\textbf{User query:}\\
\{query\}\\

Return JSON only in this exact shape:\\
\{\\
\quad "query": "\textless{}original query\textgreater{}",\\
\quad "rubrics": [\\
\quad\quad \{\\
\quad\quad\quad "dimension": "Coverage / Completeness",\\
\quad\quad\quad "title": "\textless{}short rubric title\textgreater{}",\\
\quad\quad\quad "description": "\textless{}one concise query-specific sentence\textgreater{}",\\
\quad\quad\quad "importance": "critical"\\
\quad\quad \}\\
\quad ]\\
\}\\

\textbf{Rules:}\\
- Generate exactly 5 to 7 rubric items.\\
- Use only these dimensions:\\
\quad - Coverage / Completeness\\
\quad - Depth / Insight / Explanation\\
\quad - Grounding / Citation Quality\\
\quad - Instruction / Context Alignment\\
\quad - Communication Quality\\
- Use only these importance values:\\
\quad - critical\\
\quad - important\\
\quad - nice\_to\_have\\
- Make each item specific, observable, and non-overlapping.\\
- Cover at least 3 dimensions when possible.\\
- Reflect comparisons, recommendations, trade-offs, critiques, benchmarks, or uncertainty when relevant.\\
- If the query needs evidence, citations, freshness, or benchmarks, include grounding/citation quality.\\
- Do not add extra keys.\\
- No markdown. No explanation. JSON only.
\end{promptbox}

\section{Policy Prompt}
\label{app:policy_prompt}

The following system prompt is used for RL rollouts.

\begin{promptbox}[title=Policy System Prompt]
You are a research assistant who answers questions through iterative reasoning and research.\\

Follow this protocol exactly.\\

\textbf{Tags You May Generate}\\
- \textless{}think\textgreater{}...\textless{}/think\textgreater{}\\
- \textless{}call\_tool name="google\_search"\textgreater{}\\
\quad search query\\
\quad \textless{}/call\_tool\textgreater{}\\
- \textless{}call\_tool name="browse\_webpage"\textgreater{}\\
\quad \url{https://example.com/page}\\
\quad \textless{}/call\_tool\textgreater{}\\
- \textless{}answer\textgreater{}...\textless{}/answer\textgreater{}\\

\textbf{System-Inserted Tags}\\
The system will execute your tool calls and insert the results back into the conversation using:\\
- \textless{}tool\_output\textgreater{}...\textless{}/tool\_output\textgreater{}\\
- \textless{}snippet\textgreater{}...\textless{}/snippet\textgreater{}\\
- \textless{}webpage\textgreater{}...\textless{}/webpage\textgreater{}\\

Do not generate \textless{}tool\_output\textgreater{}, \textless{}snippet\textgreater{}, or \textless{}webpage\textgreater{} yourself.\\

\textbf{Tool Rules}\\
- Use \textless{}call\_tool name="google\_search"\textgreater{}...\textless{}/call\_tool\textgreater{} for general web search. The content must be a search query, not a URL.\\
- Use \textless{}call\_tool name="browse\_webpage"\textgreater{}...\textless{}/call\_tool\textgreater{} to open a specific URL, usually one returned by a previous google search result.\\
- Use browse\_webpage only when the search snippets are not enough to answer confidently.\\

\textbf{Required Output Structure}\\
Your response must follow this structure:\\
1. Start with exactly one non-empty \textless{}think\textgreater{}...\textless{}/think\textgreater{}.\\
2. Then perform one or more research segments.\\
3. In each research segment, first do one or more google\_search cycles, and then optionally do zero or more browse\_webpage cycles.\\
Each tool cycle must follow:\\
\quad \textless{}call\_tool name="..."\textgreater{}...\textless{}/call\_tool\textgreater{}\\
\quad \textless{}tool\_output\textgreater{}...\textless{}/tool\_output\textgreater{}\\
\quad \textless{}think\textgreater{}...\textless{}/think\textgreater{}\\
4. End with exactly one final \textless{}answer\textgreater{}...\textless{}/answer\textgreater{}.\\

\textbf{Additional Constraints}\\
- Do not output any text outside these tags.\\
- Do not skip the \textless{}think\textgreater{} block before the final \textless{}answer\textgreater{}.\\
- Only output \textless{}answer\textgreater{} when you have enough information for a complete response.\\
- \textless{}answer\textgreater{}...\textless{}/answer\textgreater{} may appear only once, and it must be the final block in the entire response.\\
- The final answer must be fully contained inside \textless{}answer\textgreater{}...\textless{}/answer\textgreater{}.
\end{promptbox}

\begin{promptbox}[title=Complete Example]
Question: What is the capital of France and what is one famous museum there?\\

\textless{}think\textgreater{}I should first confirm the capital of France, then gather one well-known museum in that city before answering.\textless{}/think\textgreater{}\\

\textless{}call\_tool name="google\_search"\textgreater{}\\
\quad capital of France famous museum in Paris\\
\quad \textless{}/call\_tool\textgreater{}\\

\textless{}tool\_output\textgreater{}\\
\quad \textless{}snippet\textgreater{}\\
\quad Title: Paris - Wikipedia\\
\quad URL: \url{https://en.wikipedia.org/wiki/Paris}\\
\quad Search Snippet: Paris is the capital and most populous city of France.\\
\quad \textless{}/snippet\textgreater{}\\

\quad \textless{}snippet\textgreater{}\\
\quad Title: Louvre Museum\\
\quad URL: \url{https://www.louvre.fr/en}\\
\quad Search Snippet: The Louvre is a world-famous museum in Paris, France.\\
\quad \textless{}/snippet\textgreater{}\\
\textless{}/tool\_output\textgreater{}\\

\textless{}think\textgreater{}The search results identify a strong candidate, and I should browse the museum's official page before answering.\textless{}/think\textgreater{}\\

\textless{}call\_tool name="browse\_webpage"\textgreater{}\\
\quad \url{https://www.louvre.fr/en}\\
\quad \textless{}/call\_tool\textgreater{}\\

\textless{}tool\_output\textgreater{}\\
\quad \textless{}webpage\textgreater{}\\
\quad Title: Louvre Museum\\
\quad URL: \url{https://www.louvre.fr/en}\\
\quad Content: The Louvre is a museum in Paris and is one of the most famous museums in the world.\\
\quad \textless{}/webpage\textgreater{}\\
\textless{}/tool\_output\textgreater{}\\

\textless{}think\textgreater{}I now have enough information to answer concisely.\textless{}/think\textgreater{}\\
\textless{}answer\textgreater{}Paris is the capital of France. One famous museum there is the Louvre Museum.\textless{}/answer\textgreater{}
\end{promptbox}

\section{Implementation Details}
\label{app:implementation_details}
All RL experiments are conducted on 8 NVIDIA H800 GPUs.
The computational budget varies across tournament configurations,
ranging from approximately 1,696 to 3,648 GPU hours per training run.
We report the main training, rollout, tool-interaction, and judge hyperparameters for reproducibility. Unless otherwise specified, all reported results are from a single training run.

\begin{table}[h]
\centering
\small
\setlength{\tabcolsep}{4pt}
\begin{tabularx}{\columnwidth}{>{\bfseries}l X r}
\hline
\textbf{Group} & \textbf{Hyperparameter} & \textbf{Value} \\
\hline
\multicolumn{3}{l}{\textit{Training}} \\
& Training batch size & 16 \\
& Mini-batch size & 16 \\
& Micro-batch size per GPU & 1 \\
& Learning rate & 1e-6 \\
& Entropy coefficient & 0 \\
\hline
\multicolumn{3}{l}{\textit{Rollout}} \\
& Rollout engine & vLLM \\
& Max assistant turns per trajectory & 8 \\
& Policy rollout temperature & 1.0 \\
& Policy rollout top-$p$ & 1.0 \\
& Rollout log-prob micro-batch size per GPU & 2 \\
& Maximum response length & 18,000 tokens \\
\hline
\multicolumn{3}{l}{\textit{Tool interaction}} \\
& Maximum tool response length & 3,000 tokens\\
& Tool response truncation side & middle \\
\hline
\multicolumn{3}{l}{\textit{Judge}} \\
& Judge temperature & 0.0 \\
& Judge top-$p$ & 0.95 \\
\hline
\multicolumn{3}{l}{\textit{Validation}} \\
& Validation temperature & 0.0 \\
& Validation top-$p$ & 1.0 \\
\hline
\end{tabularx}
\caption{Main implementation hyperparameters used in our experiments.}
\label{tab:implementation_hyperparameters}
\end{table}

\section{Tournament Judge Prompt}
\label{app:tournament_judge_prompt}

For Tournament-GRPO, the prompt asks the judge to select a fixed number of winners according to the query and rubrics.

\begin{promptbox}[title=Tournament Judge Prompt]
You are an expert judge for deep-research quality.\\
Compare the candidate responses to the same query using the rubric below.\\

\textbf{Query:}\\
\{query\}\\

\textbf{Rubric (description, dimension, importance, title):}\\
\{rubric\_block\}\\

\textbf{Candidates:}\\
\{candidate\_block\}\\

\textbf{Selection rule:}\\
- Select exactly \{num\_winners\} winner(s).\\
- Prefer responses that are accurate, complete, well-supported, and directly answer the query.\\
- Return JSON only, without markdown or explanations.\\
- Use 1-based candidate numbers with this schema:\\
\quad \{"winners": [1]\}
\end{promptbox}

\section{Baseline Judge Prompts}
\label{app:baseline_judge_prompts}

\paragraph{Implicit scoring.}
The implicit scoring baseline asks the judge to evaluate all rubrics jointly and return one scalar score.

\begin{promptbox}[title=Implicit Rubric Scoring Prompt]
You are an expert scorer for deep-research quality.\\
Using the provided rubric, score the response to the query below.\\

\textbf{Query:}\\
\{query\}\\

\textbf{Response:}\\
\{rollout\_answer\}\\

\textbf{Rubric (description, dimension, importance, title):}\\
\{rubric\_block\}\\

\textbf{Scoring rule:}\\
- Return one final score in [0, 10], where 0 is worst and 10 is best.\\
- Do not output explanations.\\
- Output must be JSON only, with this schema:\\
\{\\
\quad "score": 7.5\\
\}
\end{promptbox}

\paragraph{Explicit scoring.}
The explicit scoring baseline asks the judge to evaluate exactly one rubric at a time. The final score is computed by aggregating all per-rubric scores with importance-aware weights.

\begin{promptbox}[title=Explicit Rubric Scoring Prompt]
You are an expert scorer for deep-research quality.\\
You must score the response against exactly ONE rubric only.\\
Do not aggregate across any other rubrics.\\

\textbf{Query:}\\
\{query\}\\

\textbf{Response:}\\
\{rollout\_answer\}\\

\textbf{Single rubric to score:}\\
title: \{title\}\\
dimension: \{dimension\}\\
importance: \{importance\}\\
description: \{description\}\\

\textbf{Scoring rule:}\\
- Evaluate only this single rubric.\\
- Return one final score in [0, 10], where 0 is worst and 10 is best.\\
- Do not output explanations.\\
- Output must be JSON only, with this schema:\\
\{\\
\quad "score": 7.5\\
\}
\end{promptbox}

\paragraph{Pairwise comparison.}
The pairwise comparison baseline compares two same-query rollouts at
a time. For each rollout, we compute its win rate over all pairwise comparisons
within the same-query group and use this win rate as the scalar reward.

\begin{promptbox}[title=Pairwise Comparison Prompt]
You are an expert judge for deep-research quality.\\
Compare the candidate responses to the same query using the rubric below.\\

\textbf{Query:}\\
\{query\}\\

\textbf{Rubric (description, dimension, importance, title):}\\
\{rubric\_block\}\\

\textbf{Candidates:}\\
\{candidate\_block\}\\

\textbf{Selection rule:}\\
- Select exactly 1 winner.\\
- Prefer responses that are accurate, complete, well-supported, and directly answer the query.\\
- Return JSON only, without markdown or explanations.\\
- Use 1-based candidate numbers with this schema:\\
\{\\
\quad "winners": [1]\\
\}
\end{promptbox}

\section{Baseline Reward Computation}
\label{app:baseline_reward_computation}
\paragraph{Implicit scoring.}
For implicit rubric scoring, the judge returns a raw score $z_i \in [0,10]$ for rollout $i$. We normalize it as:
\begin{equation}
    r_i^{\mathrm{imp}} = \frac{z_i}{10}.
\end{equation}
The final reward is computed as:
\begin{equation}
    r_i = r_i^{\mathrm{imp}} + r_i^{\mathrm{fmt}}.
\end{equation}

\paragraph{Explicit scoring.}
For explicit rubric scoring, each rubric is evaluated independently. Let $z_{i,k}\in[0,1]$ denote the normalized score of rollout $i$ on rubric $k$, and let $w_k$ be the raw importance weight of this rubric. We use $w_k=3$ for \texttt{critical}, $w_k=2$ for \texttt{important}, and $w_k=1$ for \texttt{nice\_to\_have}. The rubric reward is:
\begin{equation}
    r_i^{\mathrm{rubric}}
    =
    \frac{\sum_{k=1}^{R} w_k z_{i,k}}
    {\sum_{k=1}^{R} w_k}.
\end{equation}
The final explicit-scoring reward combines the rubric reward and the format reward with equal weights:
\begin{equation}
    r_i
    =
    \frac{1}{2}
    \left(
    r_i^{\mathrm{rubric}} + r_i^{\mathrm{fmt}}
    \right).
\end{equation}

\paragraph{Pairwise comparison.}
For the pairwise comparison baseline, we compare all pairs of same-query
rollouts. The reward for each rollout is its pairwise win rate:
\[
r_i^{\mathrm{pair}}=\frac{w_i}{K-1},
\]
where $w_i$ is the number of pairwise wins obtained by rollout $y_i$. The final reward is computed as:
\begin{equation}
    r_i = r_i^{\mathrm{pair}} + r_i^{\mathrm{fmt}}.
\end{equation}

\section{Additional Robustness and Generalization Results}
\label{app:additional_backbone_judge_results}

\begin{table*}[htbp]
\centering
\small
\begin{tabular}{llccccc}
\hline
\textbf{Step} & \textbf{Method} & \textbf{Readability} & \textbf{Instruct.} & \textbf{Compre.} & \textbf{Insight} & \textbf{Overall} \\
\hline
\multirow{7}{*}{600}
& Zero-shot & 30.42 & 31.77 & 29.02 & 27.56 & 29.26 \\
& GRPO explicit & 41.73 & 40.90 & 38.89 & 38.80 & 39.75 \\
& GRPO implicit & 39.77 & 40.45 & 37.23 & 36.55 & 38.07 \\
& Tournament-GRPO ($M=1$) & 42.76 & 43.23 & 41.96 & 40.98 & 42.05 \\
& Tournament-GRPO ($M=2$) & 43.32 & 44.27 & 42.80 & 41.56 & 42.81 \\
& Tournament-GRPO ($M=3$) & 44.83 & 44.65 & 43.24 & 42.35 & 43.50 \\
& Tournament-GRPO ($M=4$) & 46.59 & 46.88 & 45.99 & 44.78 & 45.86 \\
\hline
\multirow{6}{*}{1100}
& GRPO explicit & 42.98 & 42.93 & 41.25 & 40.59 & 41.67 \\
& GRPO implicit & 40.60 & 40.57 & 38.48 & 38.14 & 39.16 \\
& Tournament-GRPO ($M=1$) & 46.74 & 46.84 & 45.68 & 45.02 & 45.88 \\
& Tournament-GRPO ($M=2$) & 48.80 & 49.31 & 48.79 & 48.53 & 48.82 \\
& Tournament-GRPO ($M=3$) & 50.23 & 50.12 & 48.24 & 48.05 & 48.94 \\
& Tournament-GRPO ($M=4$) & 48.78 & 48.77 & 47.97 & 47.54 & 48.12 \\
\hline
\end{tabular}
\caption{Additional results on Deep Research Bench using Qwen2.5-3B-Instruct as the policy backbone and Qwen2.5-72B-Instruct as the LLM judge. Tournament-GRPO uses $G=4$, $K_{\mathrm{win}}=2$, and $K_{\mathrm{final}}=2$.}
\label{tab:appendix_3b_policy_qwen25_72b_judge}
\end{table*}
\begin{table*}[t]
\centering
\small
\begin{tabular}{llccccc}
\hline
\textbf{Step} & \textbf{Method} & \textbf{Readability} & \textbf{Instruct.} & \textbf{Compre.} & \textbf{Insight} & \textbf{Overall} \\
\hline
\multirow{7}{*}{600}
& Zero-shot & 26.79 & 27.52 & 22.98 & 19.76 & 23.59 \\
& GRPO explicit & 42.04 & 44.33 & 42.21 & 37.08 & 41.07 \\
& GRPO implicit & 36.69 & 38.49 & 36.07 & 31.48 & 35.28 \\
& Tournament-GRPO ($M=1$) & 45.70 & 47.27 & 47.58 & 43.41 & 45.88 \\
& Tournament-GRPO ($M=2$) & 48.93 & 50.15 & 50.47 & 48.98 & 49.77 \\
& Tournament-GRPO ($M=3$) & 48.56 & 49.80 & 50.49 & 48.40 & 49.38 \\
& Tournament-GRPO ($M=4$) & 46.22 & 47.61 & 46.27 & 42.86 & 45.47 \\
\hline
\multirow{6}{*}{1100}
& GRPO explicit & 48.47 & 49.26 & 48.17 & 45.62 & 47.65 \\
& GRPO implicit & 38.46 & 43.45 & 39.48 & 35.10 & 38.95 \\
& Tournament-GRPO ($M=1$) & 48.41 & 51.32 & 52.56 & 52.30 & 51.65 \\
& Tournament-GRPO ($M=2$) & 47.64 & 49.37 & 50.09 & 48.94 & 49.28 \\
& Tournament-GRPO ($M=3$) & 49.58 & 51.67 & 50.23 & 48.39 & 50.04 \\
& Tournament-GRPO ($M=4$) & 53.04 & 55.92 & 56.06 & 56.53 & 55.75 \\
\hline
\end{tabular}
\caption{Additional results on Deep Research Bench using Qwen2.5-7B-Instruct as the policy backbone and Qwen3-30B-A3B-Instruct-2507 as the LLM judge. Tournament-GRPO uses $G=4$, $K_{\mathrm{win}}=2$, and $K_{\mathrm{final}}=2$.}
\label{tab:appendix_7b_policy_qwen3_30b_judge}
\end{table*}

We further evaluate Tournament-GRPO under different policy backbones and judge models.  Unless otherwise specified, all experiments in this section use the same tournament setting: $G=4$, $K_{\mathrm{win}}=2$, and $K_{\mathrm{final}}=2$.

\subsection{Qwen2.5-3B-Instruct Policy with Qwen2.5-72B-Instruct Judge}

\begin{figure}[h]
\centering
\includegraphics[width=\columnwidth]{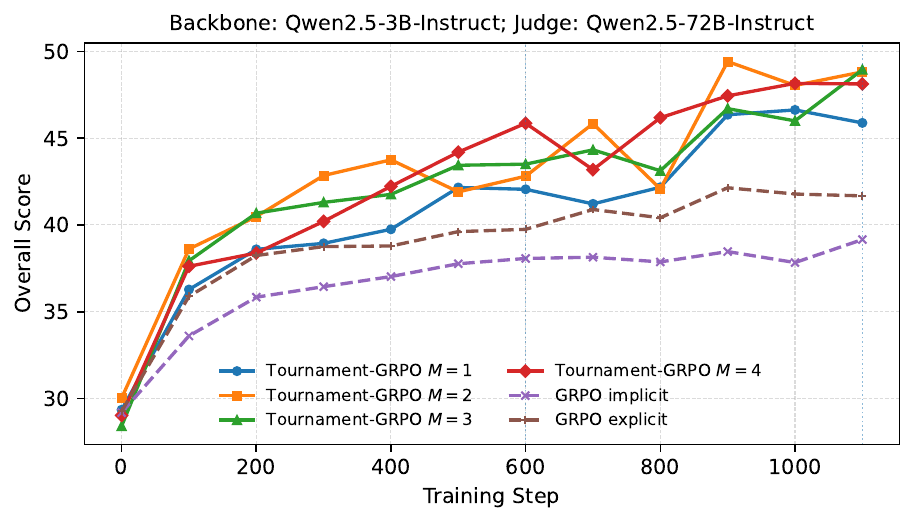}
\caption{Overall score on Deep Research Bench using Qwen2.5-3B-Instruct as the policy backbone and Qwen2.5-72B-Instruct as the LLM judge. Tournament-GRPO is evaluated with $G=4$, $K_{\mathrm{win}}=2$, and $K_{\mathrm{final}}=2$.}
\label{fig:appendix_3b_policy_qwen25_72b_judge}
\end{figure}

Table~\ref{tab:appendix_3b_policy_qwen25_72b_judge} and Figure~\ref{fig:appendix_3b_policy_qwen25_72b_judge} show the results with a smaller Qwen2.5-3B-Instruct policy backbone. Tournament-GRPO consistently outperforms GRPO implicit and GRPO explicit across both checkpoints. At step 1100, the best Tournament-GRPO setting reaches 48.94 overall, compared with 39.16 for GRPO implicit and 41.67 for GRPO explicit. This indicates that group-wise tournament rewards remain effective when the policy backbone is scaled down.

\subsection{Qwen2.5-7B-Instruct Policy with Qwen3-30B-A3B-Instruct Judge}

\begin{figure}[t]
\centering
\includegraphics[width=\columnwidth]{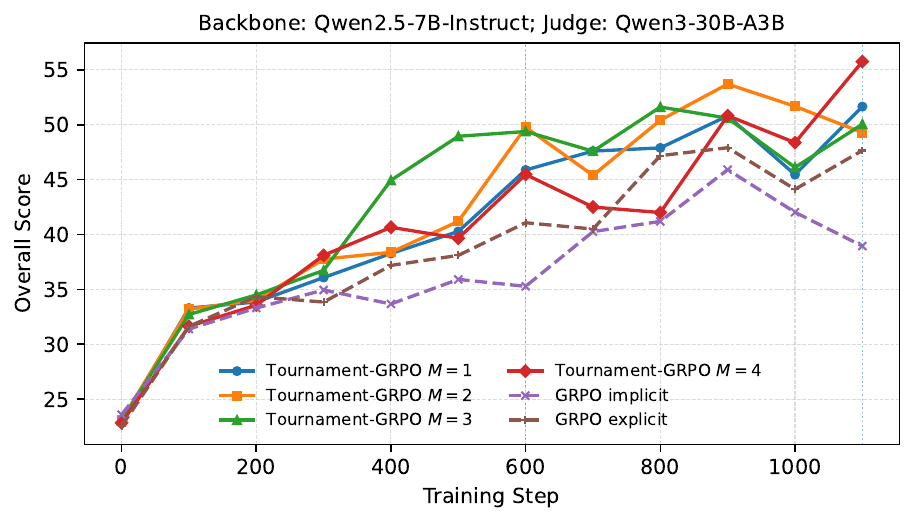}
\caption{Overall score on Deep Research Bench using Qwen2.5-7B-Instruct as the policy backbone and Qwen3-30B-A3B-Instruct-2507 as the LLM judge. Tournament-GRPO is evaluated with $G=4$, $K_{\mathrm{win}}=2$, and $K_{\mathrm{final}}=2$.}
\label{fig:appendix_7b_policy_qwen3_30b_judge}
\end{figure}

Table~\ref{tab:appendix_7b_policy_qwen3_30b_judge} and Figure~\ref{fig:appendix_7b_policy_qwen3_30b_judge} show the results when replacing the judge with Qwen3-30B-A3B-Instruct-2507. Tournament-GRPO again outperforms the GRPO baselines. At step 1100, the best Tournament-GRPO setting reaches 55.75 overall, compared with 38.95 for GRPO implicit and 47.65 for GRPO explicit. These results suggest that the effectiveness of tournament-based group-wise rewards is not tied to a single judge model.

\begin{table*}[h]
\centering
\small
\setlength{\tabcolsep}{4pt}
\begin{tabular}{lccccc}
\hline
\textbf{Method} & \textbf{Compre.} & \textbf{Insight} &
\textbf{Instruct.} & \textbf{Readability} & \textbf{Overall} \\
\hline
Zero-shot & $13.84\pm0.05$ & $12.53\pm0.05$ & $13.78\pm0.06$ & $16.95\pm0.06$ & $13.86\pm0.05$ \\
Dr. GRPO implicit & $27.45\pm0.03$ & $25.91\pm0.03$ & $29.95\pm0.04$ & $31.28\pm0.03$ & $28.08\pm0.03$ \\
Dr. GRPO explicit & $29.80\pm0.03$ & $27.87\pm0.03$ & $31.97\pm0.04$ & $33.03\pm0.02$ & $30.14\pm0.03$ \\
GDPO & $33.40\pm0.03$ & $31.17\pm0.03$ & $35.09\pm0.04$ & $36.91\pm0.02$ & $33.58\pm0.03$ \\
GRPO explicit & $26.37\pm0.07$ & $23.69\pm0.06$ & $23.69\pm0.07$ & $25.08\pm0.05$ & $24.71\pm0.06$ \\
GRPO implicit & $30.54\pm0.04$ & $27.77\pm0.04$ & $31.75\pm0.05$ & $34.32\pm0.03$ & $30.49\pm0.04$ \\
Pairwise & $36.48\pm0.05$ & $32.91\pm0.04$ & $36.48\pm0.05$ & $36.45\pm0.03$ & $35.32\pm0.04$ \\
Tournament, $M=1$ & $40.00\pm0.06$ & $36.48\pm0.06$ & $40.47\pm0.07$ & $39.72\pm0.04$ & $38.90\pm0.06$ \\
Tournament, $M=2$ & $42.65\pm0.10$ & $39.83\pm0.10$ & $43.14\pm0.10$ & $39.93\pm0.08$ & $41.45\pm0.09$ \\
Tournament, $M=3$ & $39.90\pm0.06$ & $36.02\pm0.06$ & $40.66\pm0.07$ & $40.03\pm0.04$ & $38.83\pm0.06$ \\
Tournament, $M=4$ & $\mathbf{43.73\pm0.09}$ & $\mathbf{40.55\pm0.09}$ &
$\mathbf{44.33\pm0.09}$ & $\mathbf{42.66\pm0.07}$ &
$\mathbf{42.71\pm0.08}$ \\
\hline
\end{tabular}
\caption{Statistical robustness under three repeated evaluations of the
step-600 checkpoints. Results are mean $\pm$ standard deviation, evaluated
using Qwen3-30B-A3B-Instruct-2507.}
\label{tab:statistical_robustness}
\end{table*}

\subsection{Generalization to an Unseen Judge and Robustness to Repeated Evaluation}
\label{app:statistical_robustness}

To test whether the gains merely reflect overfitting to the training judge, we
re-evaluate the step-600 checkpoints from Table~\ref{tab:epoch1_results} using
Qwen3-30B-A3B-Instruct-2507 as an evaluation judge, while all training
rewards are produced by Qwen2.5-72B-Instruct. Table~\ref{tab:unseen_judge_generalization}
reports the overall scores under this unseen judge.

\begin{table}[h]
\centering
\small
\begin{tabular}{lc}
\hline
\textbf{Method} & \textbf{Overall} \\
\hline
Zero-shot & 14.12 \\
Dr. GRPO implicit & 28.13 \\
Dr. GRPO explicit & 30.02 \\
GDPO & 33.53 \\
GRPO explicit & 24.22 \\
GRPO implicit & 30.45 \\
Exhaustive pairwise & 35.50 \\
Tournament, $M=1$ & 38.64 \\
Tournament, $M=2$ & 42.02 \\
Tournament, $M=3$ & 38.71 \\
Tournament, $M=4$ & \textbf{43.06} \\
\hline
\end{tabular}
\caption{Overall score of step-600 checkpoints under the unseen
Qwen3-30B-A3B-Instruct-2507 evaluator. Training rewards are produced by
Qwen2.5-72B-Instruct.}
\label{tab:unseen_judge_generalization}
\end{table}

All Tournament-GRPO variants outperform every baseline under the unseen judge.
Since all methods are trained with the same Qwen2.5-72B-Instruct judge but
evaluated by a different model, the gains cannot be attributed solely to
matching the judge.

To assess robustness to repeated evaluation, we further report the mean and
standard deviation over three runs using the same unseen judge in
Table~\ref{tab:statistical_robustness}.

The standard deviations are small across all metrics, and every Tournament-GRPO
configuration consistently outperforms the strongest baseline across the three
runs. These results demonstrate that the reported improvements are stable.

\subsection{Generalization beyond Deep Research Bench}
\label{app:healthbench_generalization}

We additionally evaluated our method on HealthBench-Professional under the Table~\ref{tab:epoch1_results} setting with $G=2$, $K_{\mathrm{win}}=1$, and $K_{\mathrm{final}}=1$.

\begin{table}[h]
\centering
\small
\begin{tabular}{lc}
\hline
\textbf{Method} & \textbf{Overall score} \\
\hline
Zero-shot & 47.26 \\
Dr. GRPO implicit & 49.98 \\
Dr. GRPO explicit & 49.31 \\
GDPO & 48.21 \\
GRPO explicit & 50.55 \\
GRPO implicit & 50.39 \\
Exhaustive pairwise & 51.69 \\
Tournament-GRPO, $M=1$ & 51.30 \\
Tournament-GRPO, $M=2$ & \textbf{52.54} \\
Tournament-GRPO, $M=3$ & 51.56 \\
Tournament-GRPO, $M=4$ & 51.05 \\
\hline
\end{tabular}
\caption{Results on HealthBench-Professional.}
\label{tab:healthbench_generalization}
\end{table}

Tournament-GRPO outperforms all pointwise baselines for every tested $M$, and its best configuration ($M=2$) also surpasses exhaustive pairwise comparison. These findings suggest that the gains are not specific to Deep Research Bench.

\section{Case Study}
\label{app:case_study}

\begin{table*}[t]
\centering
\small
\begin{tabular}{p{0.24\linewidth}p{0.12\linewidth}p{0.56\linewidth}}
\toprule
\textbf{Rubric} & \textbf{Importance} & \textbf{Description} \\
\midrule
Core methodology components & Critical & Specifies research design, target population, sampling strategy, data collection methods, data analysis plan, and ethical considerations tailored to low-cost housing areas in South Africa. \\
Justification of methodological choices & Critical & Justifies the chosen qualitative, quantitative, or mixed-methods approach and explains why it is appropriate for studying sanitation impacts on livelihoods. \\
Context-specific operationalization & Important & Defines and operationalizes key concepts such as poor sanitation, service delivery, livelihoods, and low-cost housing in measurable or observable terms. \\
Use of methodological literature and local evidence & Important & Draws on relevant WASH, urban poverty, and South African policy context to support design choices and acknowledges limitations. \\
Alignment with given title and scope & Critical & Keeps the methodology focused on implications of poor sanitation service delivery for livelihoods in low-cost housing areas in South Africa. \\
Validity, reliability, and bias & Important & Discusses strategies to enhance validity, reliability, or trustworthiness and to manage bias and contextual constraints. \\
Structure, clarity, and academic tone & Important & Presents the methodology in a logical, clear, and academically appropriate style suitable for a proposal or thesis chapter. \\
\bottomrule
\end{tabular}
\caption{Query-specific rubrics for the research-methodology case study.}
\label{tab:case_study_rubrics}
\end{table*}

We provide a case study from a training trajectory to illustrate how Tournament-GRPO assigns group-wise rewards by comparing same-query rollouts. This example is taken from the experiment using Qwen2.5-7B-Instruct as the policy backbone and Qwen2.5-72B-Instruct as the tournament judge, under the tournament setting $M=2$, $G=2$, $K_{\mathrm{win}}=1$, and $K_{\mathrm{final}}=1$. The trajectory is recorded at training step 451.

\paragraph{Query.}
The query asks the model to write a research methodology based on a provided title:

\begin{quote}
\small\itshape research methodology: Implications of poor sanitation service delivery on the livelihoods of people living in low cost housing areas in South Africa. write a research methodology using provided title.
\end{quote}

\paragraph{Rubrics.}
Table~\ref{tab:case_study_rubrics} shows the query-specific rubrics used for tournament judgment. The rubrics emphasize not only methodology completeness, but also methodological justification, contextual grounding, validity, bias control, and academic clarity.

\paragraph{Rollout excerpts.}
The original rollouts are long, so we show representative excerpts rather than the full trajectories. Both rollouts follow the required output format, but they receive different tournament rewards because the judge compares their rubric alignment within the same-query group.

\paragraph{High-reward rollout.}
The high-reward rollout receives the maximum normalized tournament reward. It clearly frames the research aim, motivates a mixed-methods design, and links the methodology to the South African low-cost housing context:

\begin{quote}
\small
\small\itshape This research aims to investigate the multifaceted implications of poor sanitation service delivery on the livelihoods of people living in low-cost housing areas in South Africa. Poor sanitation is a pervasive issue in these areas, which often lacks the infrastructure and services necessary for safe and sustainable living conditions, thereby affecting residents' health, economic well-being, and social stability. To achieve a comprehensive understanding of these impacts, this research will adopt a mixed-methods approach, combining qualitative and quantitative data collection and analysis methods. The study will utilize a triangulated approach to validate and complement findings, ensuring robustness and reliability of results. Ethical considerations and community engagement will be central to the research design, ensuring that the communities are informed and involved throughout the process.
\end{quote}

It also gives concrete sampling and participant-selection details:

\begin{quote}
\small\itshape
\textbf{Sampling Strategy:}
- \textbf{Purposive Sampling:} Select neighborhoods with varying levels of sanitation service provision (e.g., areas with poorly maintained pit latrines vs. areas with partially functional flush toilets).
- \textbf{Snowball Sampling:} Use initial participants to identify additional respondents who can provide deeper insights and ensure diverse perspectives.

\textbf{Participant Selection Criteria:}
- \textbf{Residence Criteria:} Households residing in low-cost housing areas with documented poor sanitation service delivery.
- \textbf{Demographic Criteria:} Participants should include residents of all ages, genders, and socioeconomic backgrounds, including children, adults, the elderly, and persons with disabilities.
- \textbf{Occupational Criteria:} Include various employment statuses to explore different livelihood impacts, such as formal sector workers, informal sector workers, and those employed in sanitation-related services.
\end{quote}

\paragraph{Low-reward rollout.}
The low-reward rollout also provides a plausible methodology, but it is less favored in tournament comparison. It begins with a broad mixed-methods description:

\begin{quote}
\small\itshape
The research will adopt a mixed-methods approach, combining quantitative and qualitative methodologies to comprehensively assess the implications of poor sanitation service delivery on the livelihoods of people living in low-cost housing areas in South Africa. This choice of approach is motivated by the need to capture both the statistical prevalence and the experiential nuances of the issue at hand. The explanatory sequential mixed-methods design will be employed, where an initial quantitative phase will provide a broad baseline and statistical understanding, and a subsequent qualitative phase will delve into the lived experiences and contextual factors that the quantitative data may not fully capture.
\end{quote}

However, compared with the high-reward rollout, it contains less consistently focused organization and some awkward or noisy phrasing in later sections. For example, the following excerpt appears in its limitations and reliability discussion:

\begin{quote}
\small
\small\itshape 
\textbf{Data Analysis and Interpretation Limitations:}
- \textbf{Generalizability:} While the chosen sampling strategy aims for generalizability, the data may still be skewed towards more accessible areas or better-educated participants, limiting the breadth of generalizable findings.
- \textbf{Complexity of Data Interpretation:} Integrating qualitative and quantitative data to draw meaningful conclusions, especially in cases where findings from different methods may not align perfectly.

\textbf{Queenship and Reliability:}
- \textbf{Qualitative Reliability:} Ensuring that qualitative findings are robust and reliable through repeated measurements, peer debriefing, and inter-rater reliability checks, though qualitative data are inherently subjective and context-specific.
- \textbf{Quantitative Reliability:} Ensuring that quantitative tools and measures are reliable through pilot testing, pretesting, and ensuring high inter-rater reliability. Use of validated scales and instruments from reputable sources will enhance reliability.
\end{quote}

This excerpt shows that the low-reward rollout is format-valid and partially relevant, but its organization and wording are less polished and less consistently aligned with the academic methodology rubrics.

\paragraph{Reward comparison.}
Table~\ref{tab:case_study_reward_comparison} compares the logged rewards for the two same-query rollouts. Both outputs receive the same format reward, but the tournament reward distinguishes their relative semantic quality.

\begin{table}[h]
\centering
\small
\setlength{\tabcolsep}{4pt}
\begin{tabular}{lcc}
\toprule
\textbf{Item} & \textbf{High} & \textbf{Low} \\
\midrule
Raw tournament score & 3.0 & 0.0 \\
Normalized tournament reward & 1.0 & 0.0 \\
Format reward & 1.0 & 1.0 \\
Final reward & 2.0 & 1.0 \\
\bottomrule
\end{tabular}
\caption{Reward comparison between high-reward and low-reward same-query rollouts.}
\label{tab:case_study_reward_comparison}
\end{table}

\paragraph{Analysis.}
The high-reward rollout receives the maximum tournament reward because it provides a more coherent methodology, gives concrete participant-selection details, and better connects the research design to sanitation service delivery and livelihoods in South African low-cost housing areas. The low-reward rollout receives no tournament reward because, although relevant and format-valid, it is less polished and includes weaker organization and noisy phrasing. This demonstrates that Tournament-GRPO provides a discriminative reward signal for open-ended writing tasks where quality depends on structure, contextual fit, and rubric-level completeness rather than exact-match correctness.

\end{document}